\RequirePackage{fix-cm}  
\documentclass[letterpaper, 10 pt, conference]{ieeeconf}  

\IEEEoverridecommandlockouts                              

\usepackage{amsmath, amsfonts, amssymb}
\usepackage{graphicx}
\usepackage{xcolor}
\usepackage{float}
\usepackage{cite}
\usepackage{bm}
\usepackage{makecell}

\usepackage{algorithm}
\usepackage{algpseudocode}

\usepackage{booktabs}
\usepackage{multirow}
\usepackage{pifont}
\newcommand{\cmark}{\textcolor{green!55!black}{\ding{51}}}
\newcommand{\xmark}{\textcolor{red!85!black}{\ding{55}}}

\usepackage{hyperref}  

\newcommand{\mat}[1]{\bm{#1}}

\graphicspath{{figures/}}

\algrenewcommand\algorithmicrequire{\textbf{Input:}}
\algrenewcommand\algorithmicensure{\textbf{Output:}}

\newif\ifanonymous
\anonymousfalse           

\ifanonymous
\title{\LARGE \bf
Humanoid\textsuperscript{*}: Open-Source Humanoid Ecosystem
for Accessible and Customizable Research and Education\thanks{\textsuperscript{*}The robot name is omitted in compliance with the IEEE RAS double-anonymous review guidelines. The linked website is likewise anonymized.}
}
\author{Anonymous Authors}
\newcommand{\robotname}{Humanoid}
\else
\title{\LARGE \bf
Arcadian: Open-Source Humanoid Ecosystem
for Accessible and Customizable Research and Education
}
\author{Zelin Shen, Gabriel Ramos Salinas, Qiyu Zhang, Yue Qin, Chen Qian, and Yanran Ding
\thanks{Authors are with the Department of Robotics, University of Michigan, Ann Arbor, MI 48109, USA.
        {\tt\small \{zelin, gramsal, qiyuzh, yueqin, chq, yanrand\}@umich.edu}}%
}
\newcommand{\robotname}{Arcadian}
\fi

\newcommand{\therobotname}{\ifanonymous the \fi\robotname}
\newcommand{\Therobotname}{\ifanonymous The \fi\robotname}

\begin{document}

\maketitle
\thispagestyle{empty}
\pagestyle{empty}

\begin{abstract}
Reproducing and modifying a meter-scale humanoid remains challenging for research and teaching groups with limited fabrication resources and engineering support.
This paper presents \robotname{}, an open-source humanoid ecosystem that offers accessible and customizable humanoid hardware and software with community support.
\Therobotname{} aims to strike a balance among cost-effectiveness, structural durability, and full customizability, lowering the barrier to entry for research and education.
Specifically, this meter-scale humanoid adopts metal-embedded 3D printing to achieve a high strength-to-cost ratio, eliminating the need for computer numerical control (CNC) machining while retaining design flexibility.
Extensive simulation and hardware tests validate the design: a reinforcement-learning (RL) locomotion policy trained in simulation transfers to the hardware, and a virtual reality (VR) teleoperation interface drives both arms for simple manipulation tasks.
The robot is around one meter tall and costs approximately \$6{,}800 to build.
All computer-aided design (CAD) models, the bill of materials, assembly documentation, and training and deployment code are publicly released, distributed through Tnkr.ai and MakerWorld and supported by a Discord server and a project forum through which community feedback informs design updates.
The ecosystem resources are available at \\
\url{https://anonymous-humanoid.gitbook.io/docs}.
\end{abstract}

\section{Introduction}

The rise of Artificial Intelligence combined with the maturity of actuators has sparked unprecedented interest in humanoid research. Commercial platforms have become affordable for a well-funded research lab, exemplified by the Unitree G1 humanoid, democratizing research in locomotion, whole-body skill learning and teleoperation~\cite{he2025asap, ze2025twist}, to name a few. Although the Unitree G1 is relatively affordable, import restrictions can limit the options available to U.S. labs, and domestic alternatives can be expensive. Even when labs can obtain a platform, proprietary hardware designs can restrict hardware modifications, and secondary development is often limited to the manufacturer’s application programming interface (API). Furthermore, the tightly integrated hardware is often treated as a black box, whose actuator, transmission, and electronics are inaccessible for modification, making research on humanoid hardware difficult.
Some manufacturers market their humanoids as open platforms that researchers can modify. K-Scale Labs, for example, publishes the mechanical and electrical designs for K-Bot along with its control software~\cite{kscalelabs_kbot}.

\begin{table*}[!t]
\centering
\caption{Physical Specifications, Fabrication Requirements, and Open-Source Resources of Humanoid Platforms}
\label{tab:platform_comparison}
\footnotesize
\setlength{\tabcolsep}{3pt}
\begin{tabular*}{\textwidth}{@{\extracolsep{\fill}}l*{10}{c}@{}}
\toprule
& \makecell[b]{MIT\\Humanoid}
& \makecell[b]{Berkeley\\Humanoid}
& \makecell[b]{HECTOR\\V2}
& \makecell[b]{Toddler-\\Bot}
& \makecell[b]{Berkeley\\Humanoid\\Lite}
& \makecell[b]{ARTEMIS}
& \makecell[b]{Duke\\Humanoid}
& \makecell[b]{AGILOped}
& \makecell[b]{MEVITA}
& \makecell[b]{\textbf{\robotname{}}\\\textbf{(ours)}} \\
& \cite{saloutos2023mit}
& \cite{liao2024berkeley}
& \cite{li2023hector, li2024pushrecovery}
& \cite{shi2025toddlerbot}
& \cite{chi2025berkeleylite}
& \cite{zhu2025artemis}
& \cite{xia2024duke}
& \cite{ficht2025agiloped}
& \cite{kawaharazuka2025mevita}
& \\
\midrule
\multicolumn{11}{@{}l}{\textit{Physical specifications}} \\
Height (m)          & $1.04$ & $0.85$ & $0.85$ & $0.56$ & $0.80$ & $1.40$ & $1.00$ & $1.10$ & ---    & $1.11$ \\
Mass (kg)           & $24$   & $16$   & $16$   & $3.4$  & $16$   & $37$   & $30$   & $14.5$ & $19.8$ & $20$   \\
Total DoFs$^{(a)}$  & 18     & 12     & 18     & 30     & 22     & 20     & 10     & 10     & 10     & 19     \\
Cost (USD)$^{(b)}$  & ---    & ---    & ---    & $6{,}000$ & $4{,}300$ & --- & $16{,}000$ & $6{,}400$ & --- & $6{,}800$ \\
\midrule
\multicolumn{11}{@{}l}{\textit{Fabrication requirements}} \\
No custom actuators & \xmark & \xmark & \cmark & \cmark & \xmark & \xmark & \cmark & \cmark & \cmark & \cmark \\
No custom CNC parts & \xmark & \xmark & \xmark & \cmark & \cmark & \xmark & \xmark & \cmark & \xmark & \cmark \\
No custom PCB       & \xmark & \xmark & ---    & \xmark & \cmark & \xmark & \xmark & \cmark & \cmark & \cmark \\
\midrule
\multicolumn{11}{@{}l}{\textit{Open-source resources and distribution}} \\
Doc                 & \xmark & \xmark & \xmark & \cmark & \cmark & \xmark & \cmark & \xmark & \xmark & \cmark \\
Code                & \xmark & \cmark & \cmark & \cmark & \cmark & \cmark & \cmark & \cmark & \cmark & \cmark \\
CAD                 & \xmark & \xmark & \xmark & \cmark & \cmark & \cmark & \cmark & \cmark & \cmark & \cmark \\
Tnkr.ai             & \xmark & \xmark & \xmark & \cmark & \cmark & \xmark & \xmark & \xmark & \xmark & \cmark \\
MakerWorld          & \xmark & \xmark & \xmark & \cmark & \cmark & \xmark & \xmark & \xmark & \xmark & \cmark \\
\midrule
\multicolumn{11}{@{}l}{\textit{Community support}} \\
Discord             & \xmark & \xmark & \xmark & \cmark & \cmark & \xmark & \xmark & \xmark & \xmark & \cmark \\
Forum               & \xmark & \xmark & \xmark & \xmark & \xmark & \xmark & \xmark & \xmark & \xmark & \cmark \\
\bottomrule
\end{tabular*}

\vspace{2pt}
\parbox{\textwidth}{\scriptsize
``---'' denotes information that is not publicly reported. \\
$^{(a)}$ Degrees of freedom actuated by motors, excluding end effectors such as grippers or hands. AGILOped drives its $12$ joints with $10$ actuators through parallel linkages. \\
$^{(b)}$ Reported build cost roughly estimated.
}
\end{table*}

Research labs have also developed their own humanoid platforms to adapt the hardware to specific research needs. Examples include the MIT Humanoid~\cite{saloutos2023mit}, Berkeley Humanoid~\cite{liao2024berkeley}, and HECTOR~\cite{li2023hector}. These platforms have supported research on dynamic locomotion and loco-manipulation. The academic publications documented the overall system architectures and key design decisions, allowing researchers from other labs to examine how the platforms are constructed and potentially reproduce the results. However, reproducing a complete robot requires detailed build documentation and a team with experience in fabrication, electronics integration, and control.

\begin{figure}[t]
    \centering
    \includegraphics[width=1\linewidth]{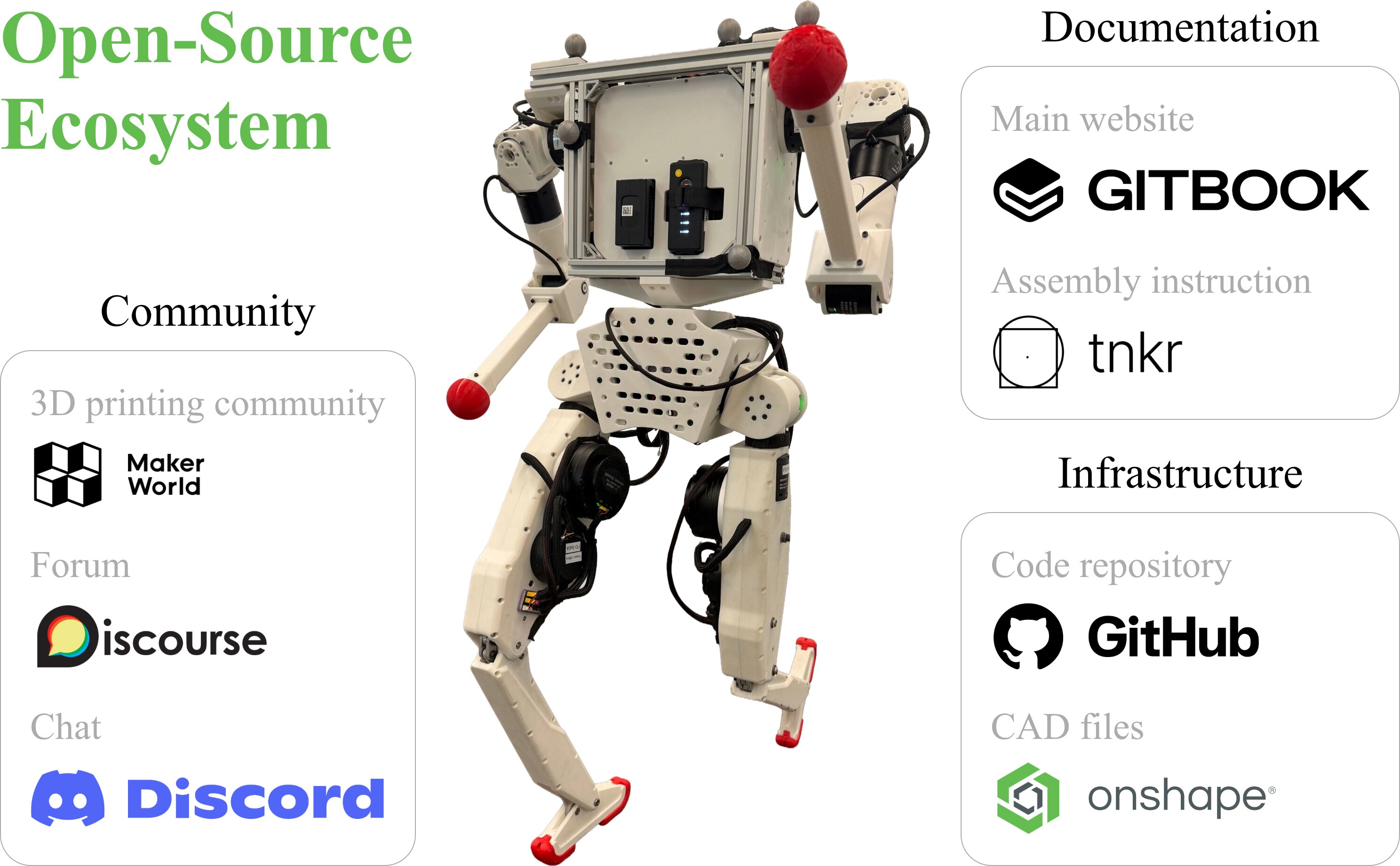}
    \caption{The \robotname{} open-source humanoid ecosystem. (Left) community support, (center) picture of the humanoid, (right) documentation and infrastructure.}
    \label{fig:ourhumanoid}
\end{figure}

Fully open-source humanoid platforms such as ToddlerBot~\cite{shi2025toddlerbot} and Berkeley Humanoid Lite~\cite{chi2025berkeleylite} released hardware designs, software, and detailed construction documentation. Both projects made print-ready designs available on MakerWorld and set up Discord servers. ToddlerBot also provided an interactive assembly guide on Tnkr.ai. Builders can fabricate replacement parts for both platforms on desktop printers.
Their compact size makes these robots easier to handle in the lab, but it can limit their ability to reach tools on a workbench designed for human operators.

Building a larger robot for these tasks also places greater loads on its leg structures. The open-source ARTEMIS~\cite{zhu2025artemis} and Duke Humanoid~\cite{xia2024duke} platforms address these demands with structural components produced by CNC machining, which provide greater structural strength and load capacity. Nevertheless, CNC machining is not always available in a laboratory or classroom, and outsourcing these components adds to the build cost and iteration time. Labs building meter-scale humanoids need an accessible, low-cost way to make load-bearing parts.

This paper presents \robotname{}, a low-cost, reproducible, open-source humanoid with an ecosystem to support its wide adoption, as shown in Fig. \ref{fig:ourhumanoid}. The robot is $1.11$~m high and can be built with a budget of around \$6{,}800. To achieve structural strength and durability while maintaining reproducibility and low cost, we adopt a composite material framework. Specifically, waterjet-cut metal is placed along the load-bearing paths, and the remaining custom structural components are produced using desktop 3D printers and can be reprinted for replacement.
The design requires no custom printed circuit boards (PCBs), custom actuators, or CNC-machined components. Beyond the waterjet-cut and 3D-printed parts, all components are commercially available.
Table~\ref{tab:platform_comparison} compares \robotname{} with representative academic humanoids in physical specifications, fabrication requirements, open-source resources, and community support.
\Therobotname{}'s CAD, bill of materials, detailed reproduction documentation, training code, and deployment code are all released as open source.
We distribute them through Tnkr.ai and MakerWorld and provide a Discord server and a project forum, aiming to close the loop between developers and users: replication issues and quick fixes reported by users flow back into the design, so that the platform continues to improve after publication.


\begin{figure*}[!t]
    \centering
    \includegraphics[width=1.0\linewidth]{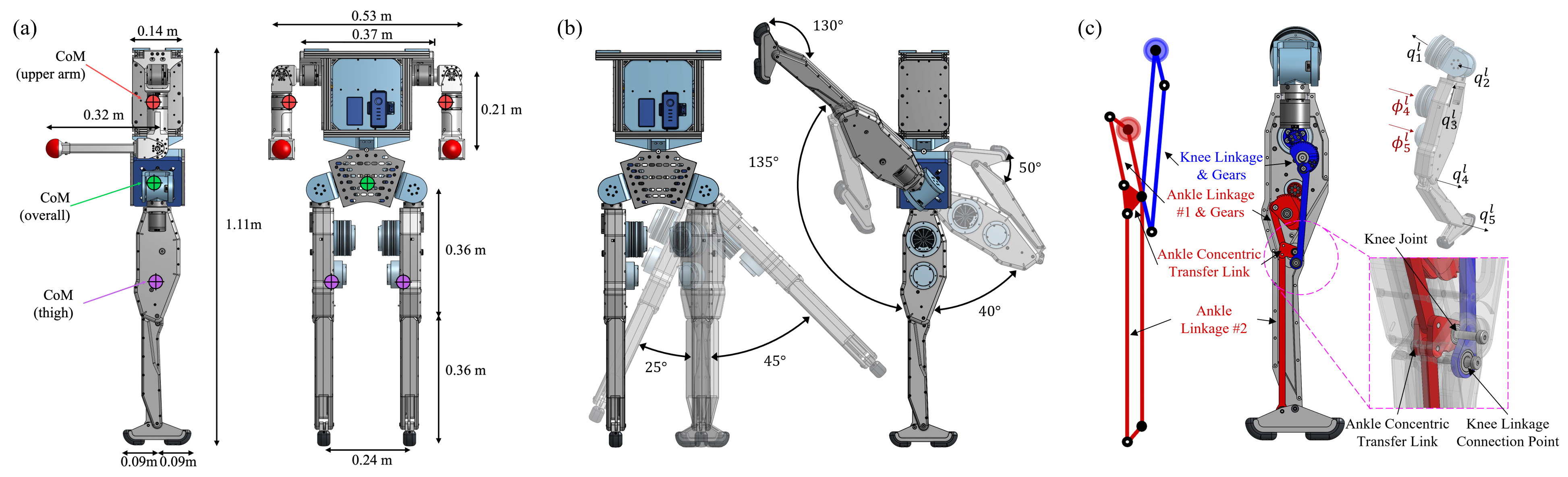}
    \caption{Mechanical overview of \robotname{}. (a) Key dimensions and center-of-mass locations. (b) Range of motion of the legs. (c) The knee (blue) and ankle-pitch (red) linkage transmissions, with the joint angles ($q_1^l$ through $q_5^l$) and actuator angles ($\phi_4^l$, $\phi_5^l$) of the left leg.}
    \label{fig:combined}
\end{figure*}

\section{Hardware Design}
\Therobotname{} follows the human morphology and stands $1.11$~m from the shoulder to the ground, with five degrees of freedom (DoF) in each leg, four in each arm, and one waist yaw, for $19$ actuated DoF in total.
All joints are driven by off-the-shelf quasi-direct drive (QDD) actuators.
The key dimensions of the robot, including its height, width, depth, and link lengths, are labeled in Fig.~\ref{fig:combined}(a).

\subsection{Leg Design}

The legs are designed to provide the range of motion necessary for dynamic movements ranging from deep squats to jumping, as well as everyday tasks such as sitting (Fig.~\ref{fig:combined}(b)). We place the hip pitch joint first in each leg's kinematic chain to avoid interference with the torso during sitting. Following human morphology, hip pitch, roll, and yaw are directly driven by motors mounted on the joints.

The knee and ankle pitch actuators are positioned above their respective joints on the thigh to reduce distal inertia for dynamic movements~\cite{chignoli2021mithumanoidrobotdesign}. In Fig.~\ref{fig:combined}(c), a series of gears and linkages accommodates this relocation and provides the power necessary for the joints.

The 3D-printed thigh and shin shells enclose the knee and ankle linkages shown in Fig.~\ref{fig:combined}(c). The shells are designed to protect these linkages from impact if the robot falls during testing.

\subsection{Metal-Embedded Shells}

\begin{figure}[htb]
    \centering
    \includegraphics[width=1.0\linewidth]{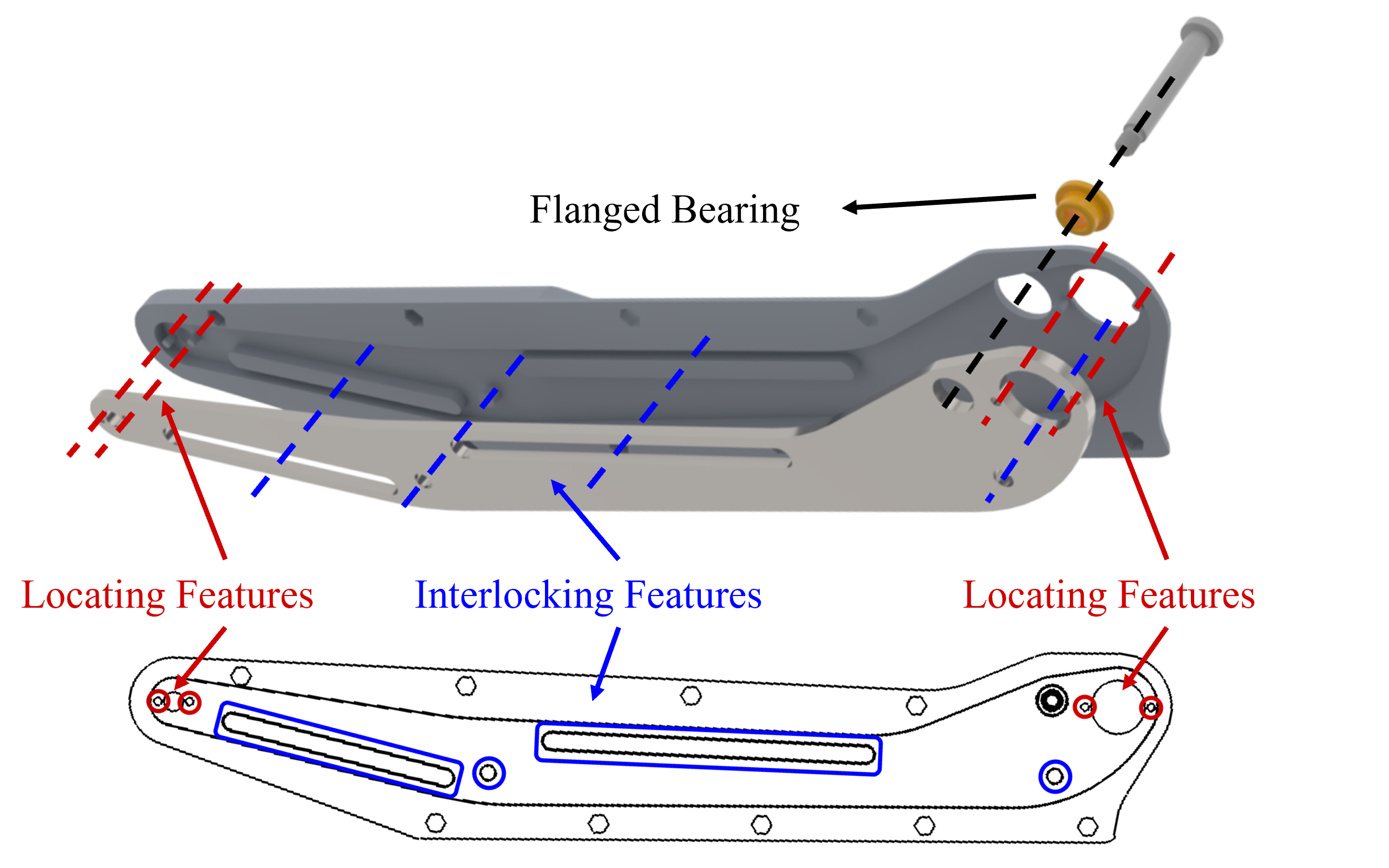}
    \caption{Metal-embedded shell. The aluminum insert is embedded within the 3D-printed shell. Interlocking features are highlighted in blue, locating features in red, and the flanged bearing is shown at the shoulder-screw interface.} 
    \label{fig:embedded}
\end{figure}

Purely 3D-printed shells are not strong enough for carrying substantial loads at the joints during locomotion, and these parts have much lower stiffness and strength than metal. To reinforce these regions, we introduce a novel composite material design where waterjet-cut metal plates are embedded in 3D-printed parts, thereby improving strength while maintaining the design flexibility. Specifically, 3.175~mm-thick 6061 aluminum alloy inserts are positioned along the primary load paths to increase the stiffness and strength of the shell, while the 3D-printed structures provide the required three-dimensional geometry and lateral support for the inserts.

\begin{figure}[htb]
    \centering
    \includegraphics[width=1.0\linewidth]{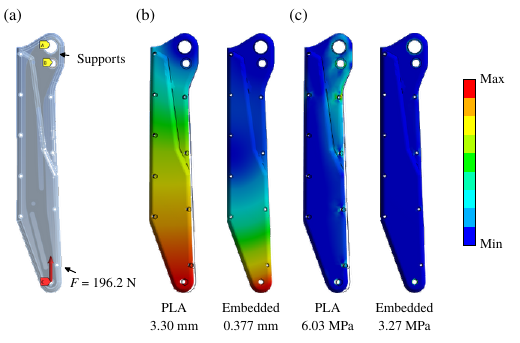}
    \caption{FEA comparison between a purely 3D-printed PLA shin shell and a metal-embedded shell under a 196.2~N static standing load. (a) Supports and applied load. (b) Total deformation. (c) von Mises stress.}
    \label{fig:fea}
\end{figure}

To quantify this improvement, we compare the metal-embedded shin shell with a purely 3D-printed PLA shell of identical geometry using finite element analysis (FEA) under a 196.2~N static standing load (Fig.~\ref{fig:fea}). The metal-embedded shell increases the equivalent vertical stiffness by a factor of 8.3, from 317.5 to 2641.6~N/mm, and reduces the peak von Mises stress in the PLA by 45.8\%, from 6.03 to 3.27~MPa.

This hybrid structure provides a high strength-to-weight ratio while maintaining low cost. Rather than CNC-machining the entire shell from metal, only the aluminum inserts are waterjet-cut, reducing fabrication cost and complexity. Embedding aluminum inserts allows the hybrid shell to carry higher loads than a purely 3D-printed structure while avoiding the mass increase associated with a fully metallic shell. Several design features and manufacturing steps are used to improve print quality, dimensional accuracy, and structural strength, as described below.

\subsubsection{Design Features}
Through-holes in the aluminum inserts mate with printed polymer bosses in the shells to form interlocking and locating features (Fig.~\ref{fig:embedded}). The interlocking features use a slip fit to ease insertion during fabrication, while the locating features use a press fit to position the inserts accurately within the shells. The shoulder screws supporting the joint linkages carry high loads, so accurate joint bores are important for minimizing radial play. Flanged bearings are installed at the shoulder-screw interfaces to limit off-axis deflection and maintain joint alignment.

\begin{table*}[!t]
    \centering
    \caption{Actuator Specifications} 
    \label{tab:actuators}
    \footnotesize
    \setlength{\tabcolsep}{4pt}
    \begin{tabular}{lcccccc p{4.6cm}}
    \toprule
    Model & Peak Torque & Weight & Max Speed & $A$ & $F_c$ & $F_v$ & Joint(s) \\
          & (Nm)        & (kg)   & (rad/s)   & (kg\,m$^2$) & (Nm) & (Nm\,s/rad) & \\
    \midrule
    DM-8009P & $40$ & $0.921$ & $17.59$ & $0.0143$ & $0.395$  & $0.0207$ & Hip pitch, knee \\
    DM-4340P & $27$ & $0.375$ & $5.45$  & $0.0246$ & $0.302$  & $0.0742$ & Hip roll/yaw, waist yaw \\
    DM-6006  & $11$ & $0.335$ & $23.46$ & $0.0017$ & $0.0659$ & $0.0073$ & Ankle pitch \\
    DM-4310  & $7$  & $0.300$ & $20.94$ & $0.0016$ & $0.0736$ & $0.0054$ & Shoulder pitch/roll/yaw, elbow pitch \\
    \bottomrule
    \addlinespace[2pt]
    \multicolumn{8}{@{}l@{}}{\scriptsize $A$, $F_c$, and $F_v$ are identified from sample units of each motor model.} \\
    \multicolumn{8}{@{}l@{}}{\scriptsize The suffix ``P'' denotes crossed-roller bearings instead of ball bearings, for higher load capacity.} \\
    \multicolumn{8}{@{}l@{}}{\scriptsize All models carry dual encoders that provide multi-turn absolute output angles.} \\
    \end{tabular}
\end{table*}

\subsubsection{Fabrication}
The aluminum inserts are first cut by waterjet. Waterjet-cut holes can exhibit dimensional inaccuracies, including through-thickness taper resulting from the characteristic kerf geometry of abrasive waterjet machining~\cite{hashish1991characteristics}. To improve dimensional accuracy, the holes are cut undersized and subsequently reamed to their final dimensions. An acrylic coating is then applied to the aluminum surface as an intermediate layer to enhance adhesion between the deposited polymer and the metal insert during printing.

Before printing, a pause command is added at the layer where the insert cavity, locating features, and interlocking features have been fully formed. Printing proceeds until this layer is reached, at which point the printer pauses and the prepared aluminum insert is placed into the cavity and aligned using the locating features. The insert is then allowed to warm on the heated print bed for several minutes before printing resumes, promoting adhesion between the subsequently deposited polymer and the coated aluminum surface. When printing resumes, the deposited polymer encapsulates the insert and interlocking features, securing the insert within the shell through both mechanical interlocking and interfacial adhesion. After printing, the flanged bearings are press-fitted into the reamed holes to complete the shoulder-screw interfaces.

\subsection{Transmission}

\subsubsection{Actuator Selection}

For accessibility and reproducibility, we build \therobotname{} around off-the-shelf QDD actuators from Damiao rather than custom in-house modules.
QDD actuators combine high torque with low reflected inertia, and their high backdrivability makes them well suited to the impact-rich dynamic motions of a humanoid~\cite{wensing2017proprioceptive}.
We selected the DM-8009P for the hip pitch and knee joints because it provides the highest peak torque among the four models in Table~\ref{tab:actuators} ($40~\mathrm{N{\cdot}m}$). The hip pitch joint is driven directly, while a 2.5:1 reduction at the knee increases the torque available from the same actuator.
Each actuator embeds its own driver and runs an independent-joint PD law with feedforward torque onboard and exchanges commands and feedback with the host computer over a flexible data-rate controller area network (FDCAN) bus.

\begin{figure}[htb]
    \centering
    \includegraphics[width=0.9\linewidth]{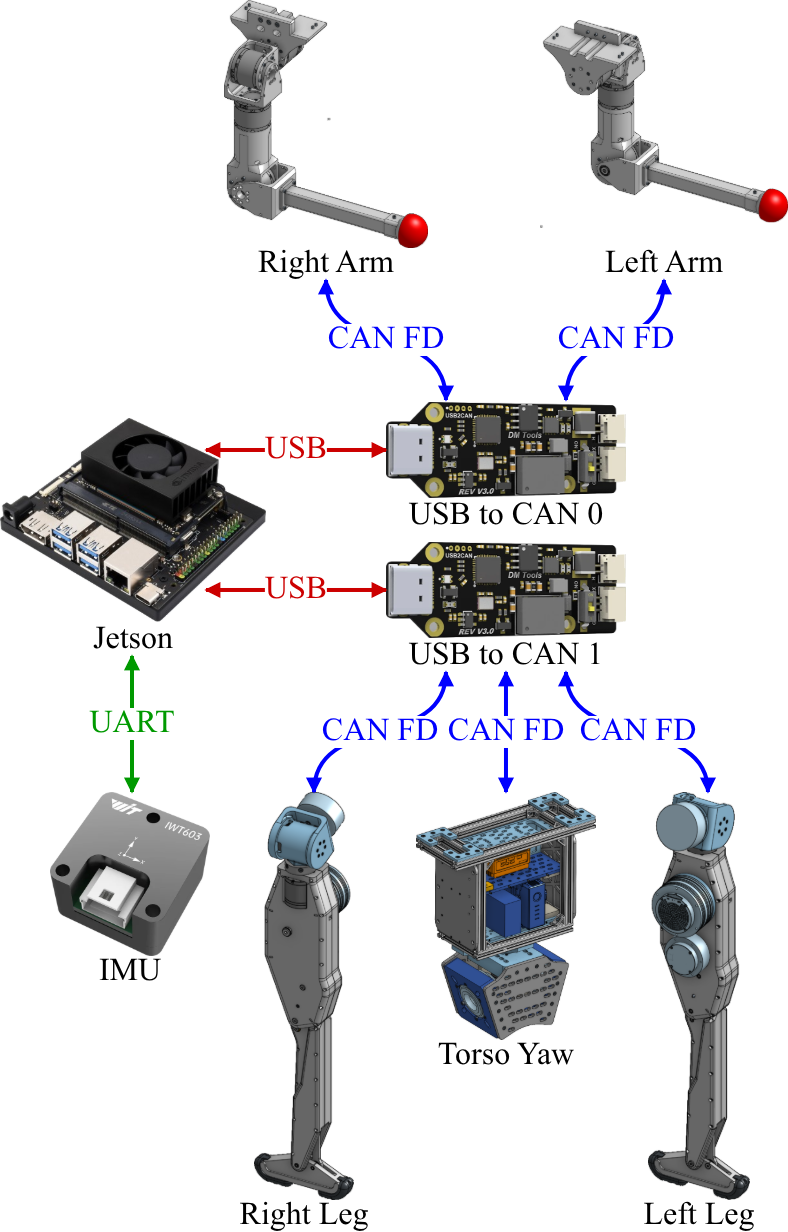}
    \caption{Simplified signal wiring of \therobotname{}}
    \label{fig:torso}
\end{figure}

\begin{figure*}[!t]
    \centering
    \includegraphics[width=1.0\linewidth]{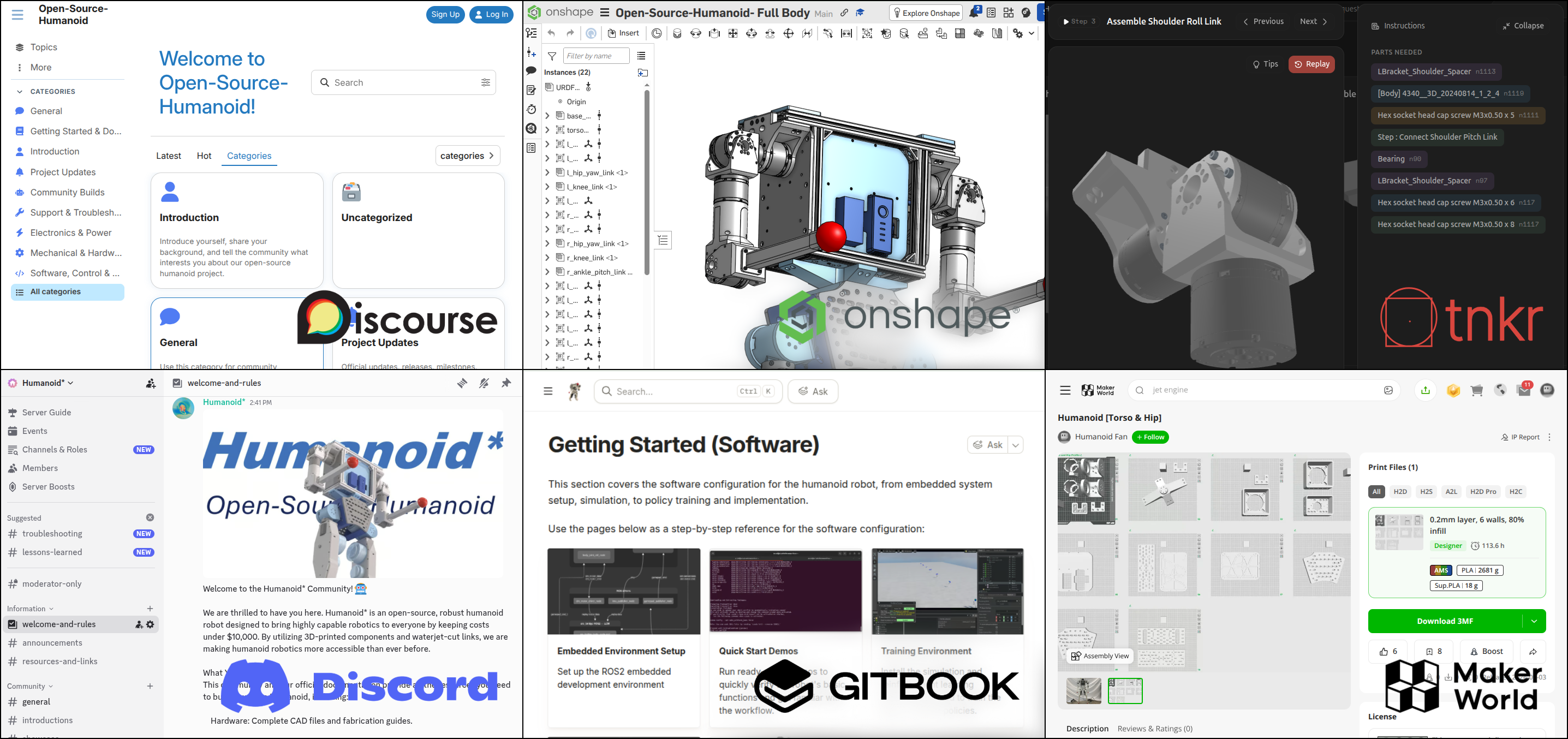}
    \caption{\Therobotname{}'s open-source community ecosystem. Top row: Discourse forum, Onshape CAD assembly, TNKR step-by-step assembly instructions. Bottom row: Discord server, GitBook documentation, MakerWorld print files.}
    \label{fig:community}
\end{figure*}

\subsubsection{Gear Design}

To power the most demanding joints, the knee and ankle pitch, 1:2.5 and 1:3 gear reductions are used, respectively. To keep costs low, the gears are 3D printed using carbon fiber-reinforced nylon 6 (PA6-CF) to withstand immense stresses generated by a maximum total tangential driving load of over 3 kN, which exceeds the material limits of standard polylactic acid (PLA). Additionally, the gears are designed with a double helix profile to distribute forces across more teeth, provide smooth actuation under dynamic shock loads and internal vibrations, and provide a naturally self-centering V-profile. The gears use a standard 1.5 module, allowing them to be easily swapped for custom CNC-machined metal alternatives if higher load capacities are required. To prevent interlaminar shear failures and increase strength, pins and screws are inserted axially into the gears to compress the layers and better distribute forces tangentially along the 3D-printed layers.

\subsubsection{Linkage Design}

The gears drive two sets of closed-chain linkages that control the knee and ankle-pitch joints. The linkages are waterjet-cut from 6 mm-thick 6061 aluminum alloy sheet to provide the necessary tensile strength to resist buckling while minimizing distal leg mass. The knee is driven by a simple 4-bar parallel linkage connected to the knee joint, enabling a range of motion from a fully extended position of $180^\circ$ to a deep squat position of $30^\circ$. 

To maintain control of the ankle at such extreme angles, the ankle is driven by a double 4-bar parallel linkage. Motion is transmitted across the knee axis $q_4^l$ via a concentric transfer link, which connects the proximal thigh linkages to the distal shin linkages. This routing prevents mechanical interference in the deep squat position, as demonstrated in Fig.~\ref{fig:combined}(c). This geometry provides the foot with its greatest range of motion ($120^\circ$) in the bent-knee position corresponding to its common locomotion range. At kinematic extremes, the ankle retains sufficient mobility to achieve full human-like extension or maintain balance in a deep squat.

\subsubsection{Leg Kinematics and Reflected Armature}

Fig.~\ref{fig:combined}(c) shows the joint and actuator coordinates on the left leg, where $q_1^l$ through $q_5^l$ are the joint angles, $\phi_4^l$ and $\phi_5^l$ are the angles of the proximally mounted knee and ankle-pitch actuators, and the superscript $l$ denotes the leg. Both legs obey the same relations.
The hip joints are driven directly, whereas the knee and ankle-pitch actuators act through the external gear stages and the parallel linkages described above. $\phi_4^l$ and $\phi_5^l$ are the output angles reported by the actuator encoders.
Closing the two linkage loops gives a constant linear map between the actuator and joint angles,
\begin{equation}
\label{eq:leg_kin}
\begin{bmatrix} q_4^l \\ q_5^l \end{bmatrix}
=
\mat{J}_p
\begin{bmatrix} \phi_4^l \\ \phi_5^l \end{bmatrix},
\qquad
\mat{J}_p =
\begin{bmatrix} 1/N_4 & 0 \\ -1/N_4 & 1/N_5 \end{bmatrix},
\end{equation}
where $N_4 = 2.5$ and $N_5 = 3$ are the external gear ratios of the knee and ankle stages, and the subscript $p$ marks the parallel transmission.
The diagonal entries are the two gear reductions, and the off-diagonal entry accounts for the ankle joint linkages passing through the knee joint, so that knee motion alone already changes the ankle joint angle.
Since $\mat{J}_p$ is constant, the velocities transform through the same matrix, and the principle of virtual work gives the torque map,
\begin{equation}
\begin{aligned}
\begin{bmatrix} \dot{q}_4^l \\ \dot{q}_5^l \end{bmatrix}
&=
\mat{J}_p
\begin{bmatrix} \dot{\phi}_4^l \\ \dot{\phi}_5^l \end{bmatrix}, \\[0.3em]
\begin{bmatrix} \tau_{q_4}^l \\[0.35em] \tau_{q_5}^l \end{bmatrix}
&=
\mat{J}_p^{-\top}
\begin{bmatrix} \tau_{\phi_4}^l \\[0.35em] \tau_{\phi_5}^l \end{bmatrix},
\qquad
\mat{J}_p^{-\top} =
\begin{bmatrix} N_4 & N_5 \\ 0 & N_5 \end{bmatrix},
\end{aligned}
\end{equation}
where $\tau_{q_i}$ and $\tau_{\phi_i}$ denote the joint and actuator torques, respectively.
The first column of $\mat{J}_p^{-\top}$ shows the knee stage amplifying its actuator torque by $N_4$, and the second column shows the ankle actuator acting on both joints through its linkage.
We apply this mapping consistently across simulation and deployment.

The closed-chain transmission also couples the actuator armatures $\mat{A}_\phi = \operatorname{diag}(A_{8009P},\, A_{6006})$ into joint space,
\begin{equation}
\label{eq:armature}
\mat{A}_q
=
\mat{J}_p^{-\top}\mat{A}_\phi\,\mat{J}_p^{-1}
=
\begin{bmatrix}
A_{q_4} & A_{q_5} \\[0.2em]
A_{q_5} & A_{q_5}
\end{bmatrix},
\end{equation}
and $A_{8009P}$ and $A_{6006}$ are the knee and ankle-pitch actuator armatures from Table~\ref{tab:actuators}.
The robot description files (URDF~\cite{tola2024urdf}/XML~\cite{todorov2012mujoco}) used for training and simulation model the leg as open joints with one scalar armature per joint, so we keep the diagonal terms $A_{q_4}$ and $A_{q_5}$.
The directly driven joints use their motor armatures unchanged.

\subsection{Electrical System}

The electrical system of \therobotname{} is entirely self-contained within the upper torso. To safeguard these critical components, the upper torso frame is constructed from standard $20$~mm aluminum T-slot extrusion. This design provides a robust, protective enclosure for the electronics while remaining easy to assemble and extend with additional hardware.

The system is powered by two DJI TB48S $24$~V batteries connected in parallel. These route through a central power distribution board that supplies the computing, sensing, and actuation subsystems. A wireless relay is integrated into this power pathway, serving as a dedicated switch to safely cut or enable the power supply to all motors during operation.

High-level computation is handled by an NVIDIA Jetson Orin NX onboard computer, while an IWT603 inertial measurement unit (IMU) provides orientation information.
Communication between the onboard computer and the distributed joint actuators is bridged by two Damiao USB2CANFD adapters, one serving the arms and the other serving the legs and waist yaw, as shown in Fig.~\ref{fig:torso}.

\section{Software and Open-Source Ecosystem}

The software stack is built on ROS~2 as a set of single-purpose nodes that communicate through typed messages: driver nodes for the Damiao actuators (over USB2CANFD), the IMU, the gamepad, and the VR headset, and a reinforcement-learning inference node as the controller.
The whole stack can be self-contained on the onboard computer, with no low-level microcontroller and no custom firmware.
Since nodes interact only through messages, swapping the motor and IMU drivers for a MuJoCo~\cite{todorov2012mujoco} simulation module allows the same controller to run in simulation.
The ROS~2 message mechanism also supports off-board operation over Ethernet.

The released resources cover the whole reproduction workflow, hosted on the platforms shown in Fig.~\ref{fig:community}.
The full CAD assembly is published on Onshape, where users can inspect the design in a browser and make their own copies for modification.
The 3D-printed parts are distributed on MakerWorld as print profiles with tuned slicer settings, so each part can be printed directly.
An interactive guide on Tnkr.ai then walks through the assembly step by step, listing the parts and fasteners needed at each stage in the correct assembly order.
The entire software stack and the training and deployment code are hosted on GitHub and work out of the box. Once the development environment is set up, users can drive the robot in simulation right away, and the included demos run directly on the hardware.
The GitBook documentation organizes all of this into one place, with hardware chapters covering the bill of materials, tools, 3D printing, assembly, and waterjet metal manufacturing, and software chapters covering environment setup, quick-start demos, training, and sim-to-real deployment.

Community support runs on two channels.
On the Discord server, builders can ask questions, post quick fixes, share their builds, and follow announcements in dedicated troubleshooting and lessons-learned channels.
The Discourse forum provides a searchable record of discussions on electronics, mechanics, and software.
Feedback gathered through both channels flows back into the released files and documentation, and we maintain the platform through this loop with continued updates and support.

\section{Experiment}

\subsection{Actuator System Identification}
\label{sec:actuator_identification}

\begin{figure}[htb]
    \centering
    \includegraphics[width=\linewidth]{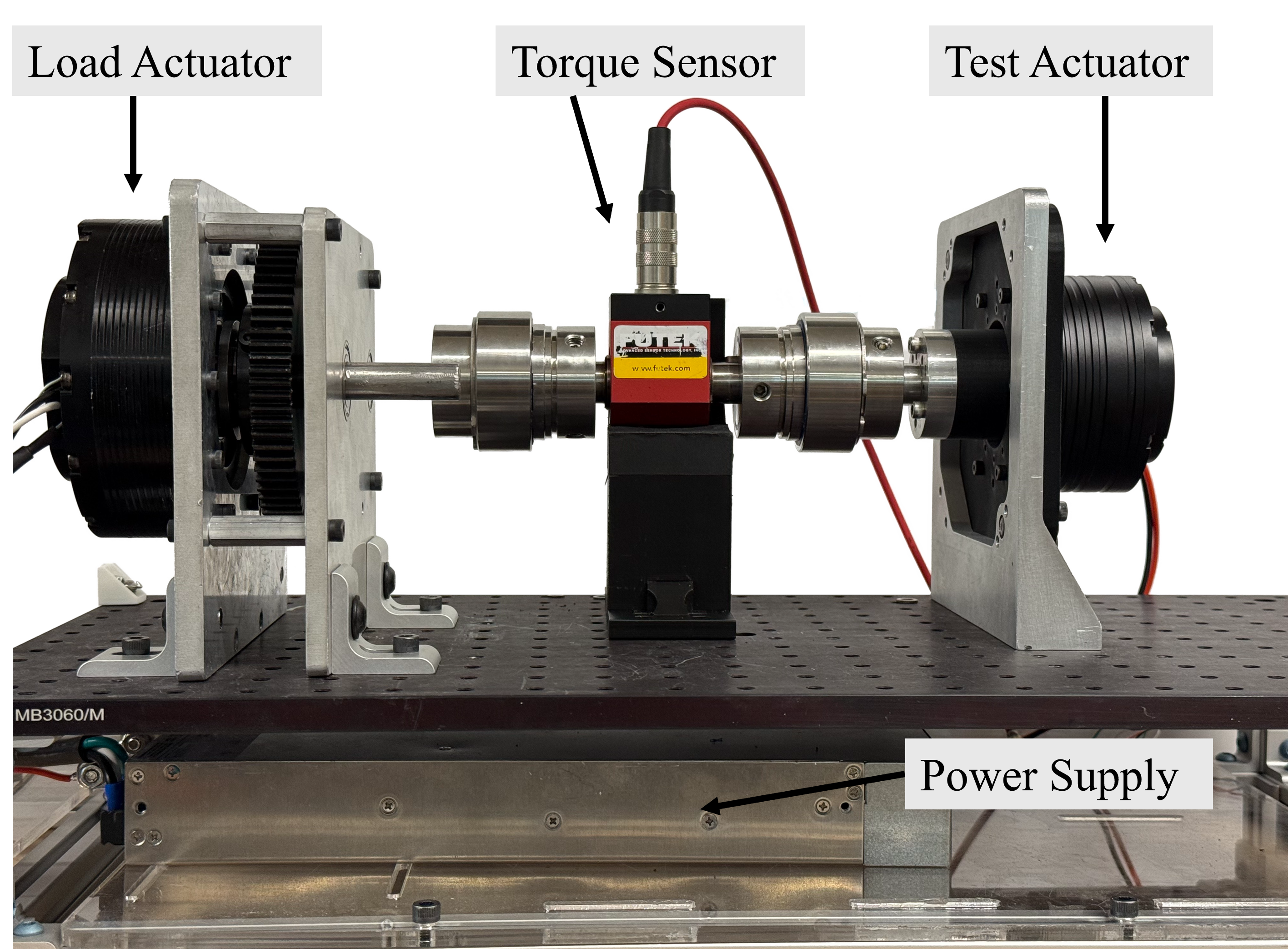}
    \caption{Dynamometer setup for actuator identification.}
    \label{fig:dyno}
\end{figure}

We identify the parameters of all four motor types using the dynamometer in Fig.~\ref{fig:dyno}, which couples the test actuator to a load motor through an in-line torque sensor.
In the loaded test, the test actuator is torque-controlled against the velocity-holding load motor, and the measured-to-reported torque ratio gives the effective torque constant.
In the no-load test, fitting the actuator's equation of motion gives the reflected armature $A$, the Coulomb friction $F_c$, and the viscous friction $F_v$.
Table~\ref{tab:actuators} lists $A$, $F_c$, and $F_v$ for each motor model.

\subsection{Setup}

The velocity-tracking locomotion controller is a reinforcement-learning policy trained in Isaac Lab~\cite{mittal2025isaac_lab} with PPO~\cite{schulman2017ppo}.
The only addition to the standard formulation is that the actor also observes the base linear velocity and base height. In training, they are taken from the simulator as ground truth, and in deployment, they are provided by a Vicon motion capture (MoCap) system, while the base orientation and angular velocity still come from the onboard IMU.
The policy runs at $50$~Hz and outputs position targets for the $11$ lower-body actuators, which match the real robot's configuration.

The upper-body teleoperation uses a Meta Quest 3S headset.
The operator's controller poses are streamed as end-effector targets for the two arms.
The targets are tracked by velocity-level differential inverse kinematics, formulated as a quadratic program with joint-limit constraints and solved by ProxQP~\cite{bambade2025proxqp}, and the resulting joint position commands are sent with a gravity-compensation feedforward torque computed from the arm URDF model using Pinocchio~\cite{carpentier2019pinocchio}.

Both controllers are first verified in the MuJoCo simulator before deployment on the robot.

\subsection{Simulated Loco-Manipulation}

\begin{figure}[htb]
    \centering
    \includegraphics[width=1.00\linewidth]{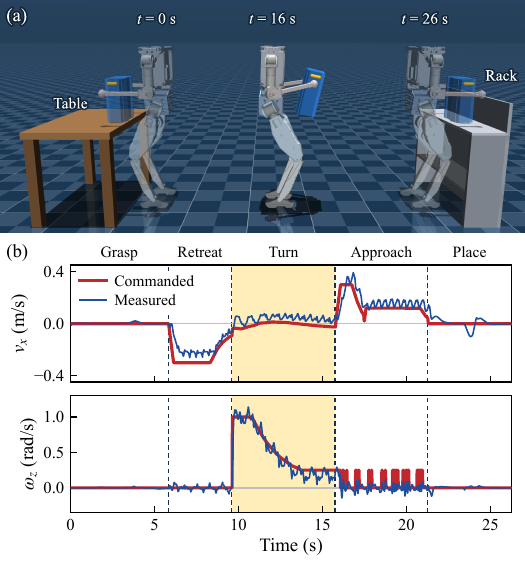}
    \caption{(a) Table-to-rack box transfer in MuJoCo. (b) Commanded and measured forward velocity and yaw rate.}
    \label{fig:sim_locomanip}
\end{figure}

In MuJoCo, the robot moves a box $2.3$~m from a table to a rack (Fig.~\ref{fig:sim_locomanip}(a)).
Following a scripted arm trajectory, it grips the roughly $1$~kg box between its forearms and lifts it off the table.
The robot then backs up, turns $180^\circ$ in place, and walks to the rack.
A waypoint controller uses the base pose to generate velocity commands for the lower-body locomotion policy.

The robot sets the box down about $26$~s after the start.
Over this interval, the root-mean-square tracking errors are $0.05$~m/s for forward velocity and $0.06$~rad/s for yaw rate (Fig.~\ref{fig:sim_locomanip}(b)).
The brief yaw-rate commands during the approach period correct the robot's heading.

The robot placed the box on the rack in all $60$ simulated trials covering different box masses ($0.64$, $1.14$, $1.64$, and $2.14$~kg), initial headings ($0^\circ$, $\pm6^\circ$, and $\pm12^\circ$), and lateral offsets ($0$ and $\pm2$~cm).
The simulation suggests that a human operator could also direct the physical robot to move a box from a table to a rack by steering it and teleoperating its arms.

\subsection{Locomotion}

\begin{figure}[htb]
    \centering
    \includegraphics[width=1.00\linewidth]{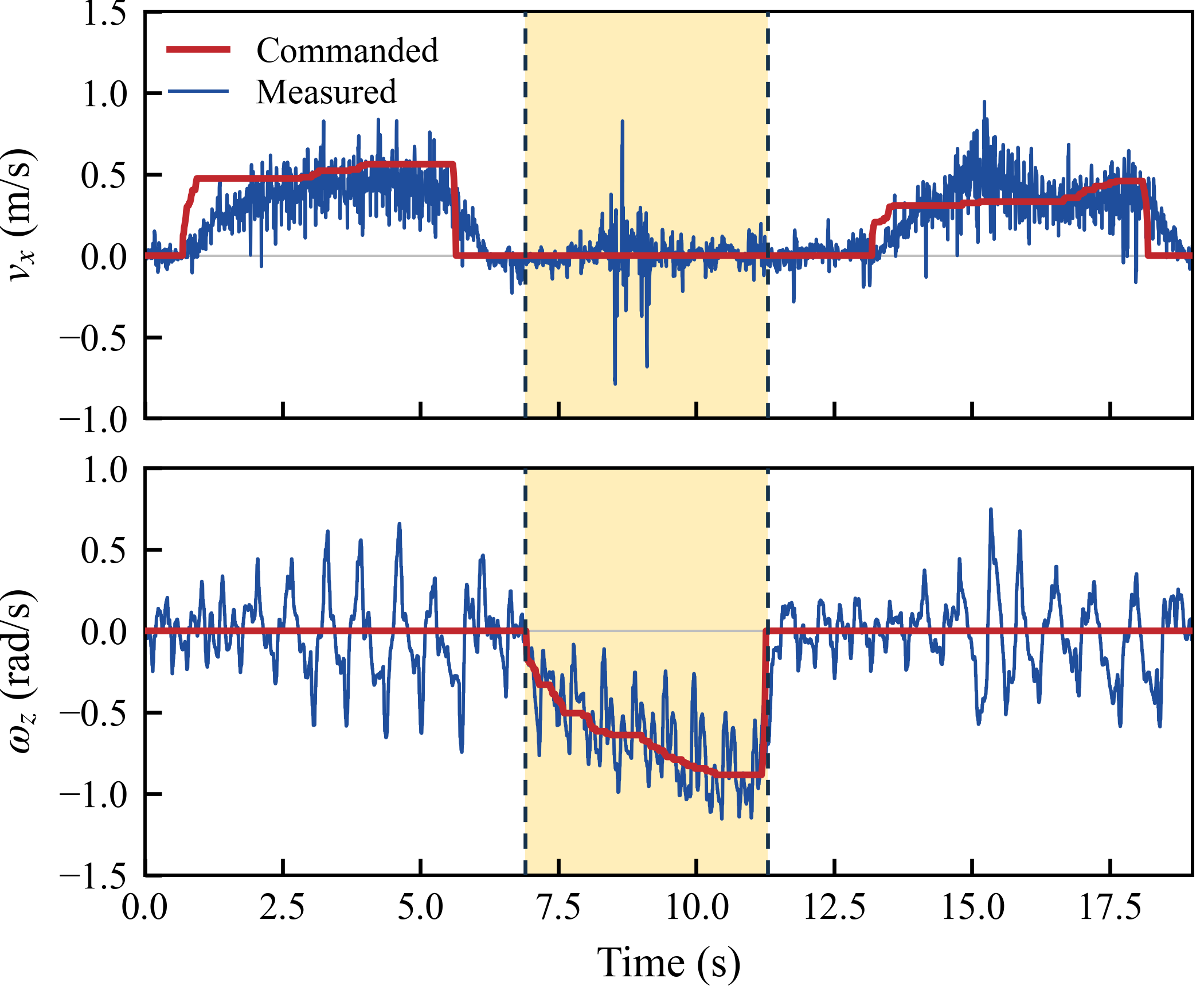}
    \caption{Commanded and measured forward velocity $v_x$ (top) and yaw rate $\omega_z$ (bottom). Commands come from the gamepad; $v_x$ is measured by MoCap and $\omega_z$ by the onboard IMU. The shaded region marks the in-place $180^\circ$ turn.}
    \label{fig:locomotion}
\end{figure}

To evaluate the sim-to-real transfer of the locomotion policy, we deployed it on the hardware and commanded the robot with a gamepad inside a MoCap arena.
In the experiment, the robot walked forward across the arena, turned $180^\circ$ in place, and walked back to its starting point.
Fig.~\ref{fig:locomotion} compares the commanded and measured forward velocity $v_x$ and yaw rate $\omega_z$.
The robot completed the round trip and followed both commands throughout the forward, turning, and return segments.
These preliminary results indicate that the locomotion policy transfers to the hardware, although the walking is not yet perfect.

\subsection{Teleoperation}

\begin{figure}[htb]
    \centering
    \includegraphics[width=1.00\linewidth]{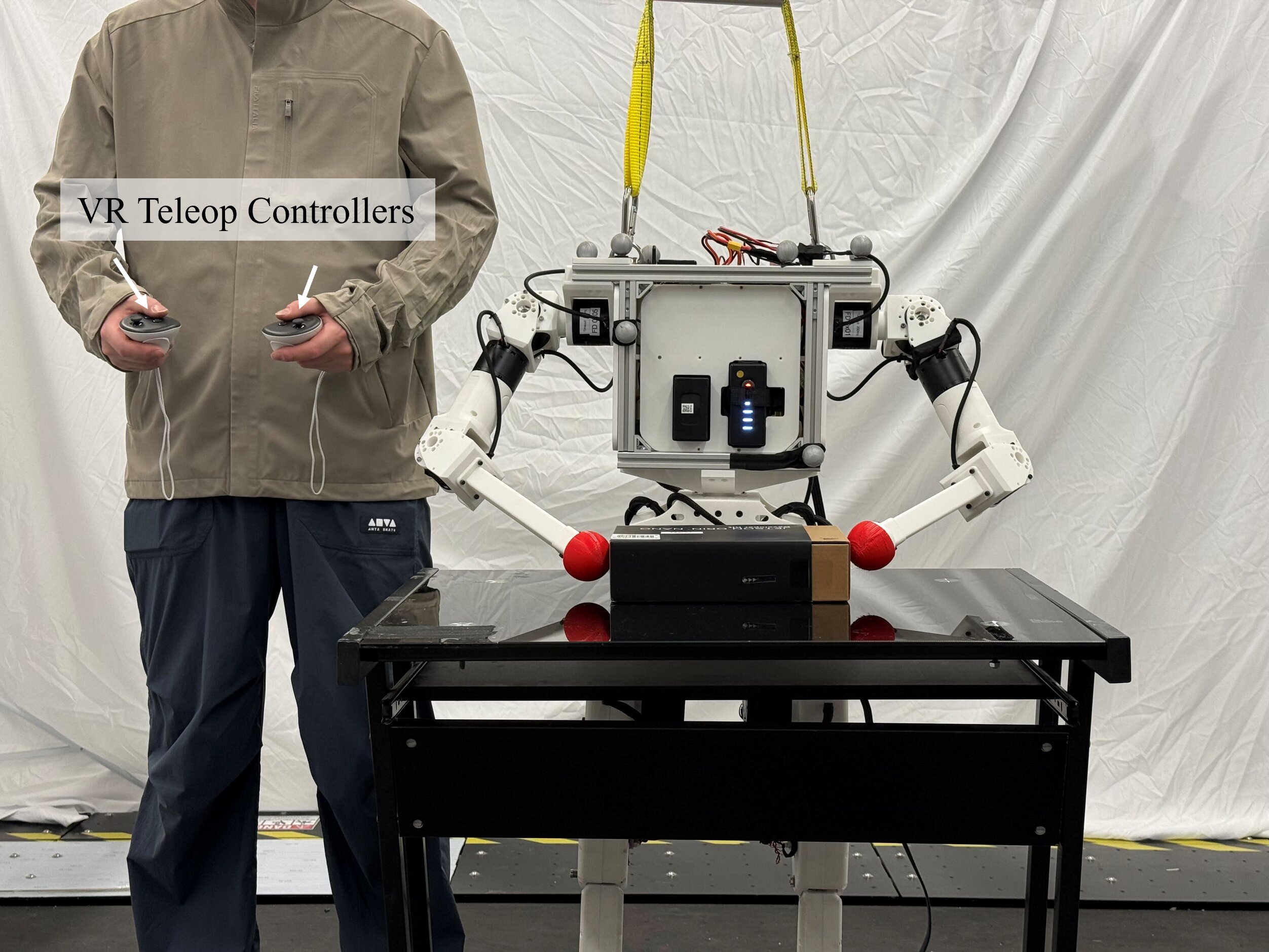}
    \caption{VR teleoperation. The operator commands the two arms through the Meta Quest 3S controllers, and the robot tracks the hand motions in real time.}
    \label{fig:teleop}
\end{figure}

An operator commands both arms with the handheld VR teleop controllers, and the robot tracks the hand motions in real time to perform simple box-pushing manipulation tasks, as shown in Fig.~\ref{fig:teleop}.

\section{Conclusion}

This work presented \robotname{}, an open-source humanoid ecosystem that spans hardware, software, and community support.
The robot is a $1.11$~m, $20$~kg, $19$-DoF platform built for approximately \$6{,}800 from off-the-shelf QDD actuators and metal-embedded 3D-printed structures, without CNC machining, custom actuators, or custom PCBs.
We demonstrated reinforcement-learning locomotion and VR-based upper-body teleoperation on the hardware.
The complete reproduction resources are released through Tnkr.ai, MakerWorld, and GitBook, with user feedback from a Discord server and a project forum informing design updates.

Our long-term goal is for \robotname{} to serve as a shared platform for humanoid research and education, maintained together with the community that rebuilds it.
The next hardware revisions will add wrists and hands to the arms.
The VR teleoperation then becomes a demonstration-collection interface for imitation learning, in the style of recent vision-language-action policies such as $\pi_0$~\cite{black2024pi0}, moving the platform from teleoperated manipulation toward autonomous loco-manipulation.
The results presented here are the first steps in that direction.










\bibliographystyle{IEEEtran}
\bibliography{reference}

\end{document}